# Experimentation on the motion of an obstacle avoiding robot

Rakhmanov Ochilbek
Dept. of Computer Science
Nile University of Nigeria
Abuja, Nigeria
ochilbek.rakhmanov@nileuniversity.edu.ng

Nzurumike Obianuju
Dept. of Computer Science
Nile University of Nigeria
Abuja, Nigeria
nzurumikeuju@yahoo.com

Amina Sani
Dept. of Computer Science
Nile University of Nigeria
Abuja, Nigeria
amina3sani@gmail.com

Rukayya Umar
Dept. of Computer Science
Nile University of Nigeria
Abuja, Nigeria
umarrukayya114@gmail.com

*Abstract*— An intelligent robot can be used for applications where a human is at significant risk (like nuclear, space, military), the economics or menial nature of the application result in inefficient use of human workers (service industry, agriculture), for humanitarian uses where there is great risk (demining an area of land mines, urban search and rescue). This paper implements an experiment on one of important fields of AI – Searching Algorithms, to find shortest possible solution by searching the produced tree. We will concentrate on Hill climbing algorithm, which is one of simplest searching algorithms in AI. This algorithm is one of most suitable searching methods to help expert system to make decision at every state, at every node. The experimental robot will traverse the maze by using sensors plugged on it. The robot used is EV3 Lego Mind-storms[15], with native software for programming LabView[15]. The reason we chose this robot is that it interacts quickly with sensors and can be reconstructed in many ways. This programmed robot will calculate the best possibilities to find way out of maze. The maze is made of wood, and it is adjustable, as robot should be able to leave the maze in any design.

*Keywords— Artificial Intelligence, Expert Systems, Robot, Obstacle Avoidance, Hill-Climbing Algorithm, Sensors*

## I. Introduction

As technology advances and smart devices become commonplace, Artificial Intelligence (AI) is becoming one of the most sought-after and most researched fields. Expert systems and smart machines are appearing in everyday products, with a variety of functionalities.

This paper implements an experiment on one of important fields of AI – Searching Algorithms, to find shortest possible solution by searching the produced tree. We will concentrate on Hill climbing algorithm, which is one of simplest searching algorithms in AI. This algorithm is one of most suitable searching methods to help expert system to make decision at every state, at every node. The experimental robot constructed and programmed by us, will traverse the maze by using sensors plugged on it.

The term Robot generally means a machine possessing human characteristics and being able to carry out the expected task programmed it to perform. Technically put by Rubin Murphy in his text book "an intelligent robot is mechanical creature which can function autonomously" and robotics is a branch of technology that deals with the design, construction, operation and application of robots [6]. A robot function based on the algorithm used in programming it that is, performs the task assigned it to perform. The choice of algorithm is completely dependent on the type of function needed; lots of algorithms exist for different purposes like Binary search algorithm. A searching algorithm is widely used algorithm for route finding.

Most people think of robots in humanlike terms communicating and doing things like people would. However, these specific subsets of robots are actually not very common. A robot can be defined as a mechanical device that is capable of performing a variety of tasks on command or according to instructions programmed in advance [1].

With the forces of nature and the unpredictability of humanity at work, the world is in constant turmoil. The full power of nature can result in earthquakes, tsunamis and devastating storms. Humanity adds to this through war, terrorist attacks and unfortunate constant turmoil. These incidents require people to search for survivors and help remove them from the site of the incident. [2].

To search and locate survivors rescuers use various pieces of equipment (sounds poles, infrared cameras and sonars) as well as dogs. To reduce the risks to the rescuer, the search is carried out slowly and delicately but this has a direct impact on the time to locate survivors. The concept of the use of a robot to act for humans naturally lends itself to the area of searching hazardous or large environments in place of or sup-porting human searchers [2].

An Obstacle Avoiding Robot is an intelligent robot, which can automatically sense and overcome obstacles on its path. Obstacle Avoidance Robot has a vast field of application. They can be used as services robots, for the purpose of household work and so many other indoor applications, they also have great importance in scientific exploration and emergency rescue, there may be places that are dangerous for humans or even impossible for humans to reach directly, then we should use robots to help us. In those challenging environments, the robots need to gather information about their surroundings to avoid obstacles [5].

AI involves using methods based on the intelligent behavior of humans and other animals to solve complex problems [6]. In other words, it is a way of making a computer, a computer-controlled robot, or a software think intelligently, in the same manner that human think.

Hill Climbing is a heuristic search used for mathematical optimization problems in the field of Artificial Intelligence.

Given a large set of inputs and a good heuristic function, it tries to find a sufficiently good solution to the problem. [3]

The robot we will use is EV3 Lego Mind-storms [4, 14]. The reason we chose this robot is that it interacts quickly with sensors and can be reconstructed in many ways.

To program the robot, we will use LEGO MIND-STORMS Education EV3 1.4.0 software [4, 14].

This programmed robot will calculate the best possibilities to find way out of maze. The maze is made of wood, and it is adjustable, as robot should be able to leave the maze in any design.

There are many experiments and researches done in this field. For instance, El-Aawar et al [9], used both of the concepts we used in this experiment, Hill climbing search algorithm and obstacle avoiding, with difference that they conducted experiment through simulation, while our experiment is real life trial.

## II. REVIEW OF RELATED LITERATURE

Since this work focuses on two aspects: (i) designing the robot and (ii) programing it using a search algorithm, our literature review focuses on the above mentioned aspects, LEGO robots and Hill Climbing searching algorithm.

Lego Robots has been used in different domain for different purposes like teaching, Stuikys et al., used in the teaching to increase students' engagement in learning [7]. Another work was done on disabled children where Lego robot was used in ascertain whether low costs robots provide a means by which disabled children can exhibit understanding of cognitive concepts [8].

Hill climbing algorithm is a local search algorithm that seeks to find optimal path from a starting position to a goal, it starts with a stochastic solution and climbs the hill by accepting better connecting solution than the one at hand the algorithm stops when it gets to either global maximum in the case of maximization or global minimum for Minimization [9,10]. Ben Cuppin gave simple description for this algorithm; hill climbing is an example of an informed search method because it uses information about the search space to search in a reasonably efficient manner. If you try to climb a mountain in fog with an altimeter but no map, you might use the hill climbing to Generate and Test approach [11].

El-Aawar and Husseini used Hill climbing search algorithm in their work to implement computer simulated artificial intelligent agent that moves in 2D and 3D environments, a set of algorithms that helps the agent finds shortest path to a known target location while avoiding all obstacles. In addressing hill-climbing limitations, map generation algorithm was introduced which assigns costs to every location in the map. They concluded by stating that the algorithm has produced a complete and logical solution for the agent in 2D and 3D environments [9].

Dash and Mishra proposed an algorithm for path planning of Robot with regards to the existing problem of path planning of Robot in the area of robotics. Their algorithm uses the Neural Network training technique. The output resulted in positive impact which is finding the next position of the robot is fast because of the neural network Technique and use of sensor system. The conclusion indicated implantation of the system is costly [12].

Another work was carried out by Liang with regards the existing path planning of robot problem. Proposed a path planning algorithm in which A* search was to calculate the grid from starting to goal as opposed to [12] above in where they used map planning algorithm to assign the cost every location in the map [13].

Over the last few years, several other researchers have presented variety of approaches to solve the maze in different ways, using different methods and different sensors

Van Putten in his Bachelor final project explained a maze solving robot that works more on Line Following category [9].

Moreover, Paul Kevan, in 2003, Leeds University, in his Final year project described a robot, designed to solve the maze with touch sensor [10].

## III. EXPERIMENTATION ON THE ROBOT

This section is presented in 4 parts; compiling the Lego robot, experimenting with Hill climbing algorithm, EV3 programming software and maze construction.

### 3.1 Assembly of the robot

All necessary parts of EV3 are listed in Figure 1 [14].

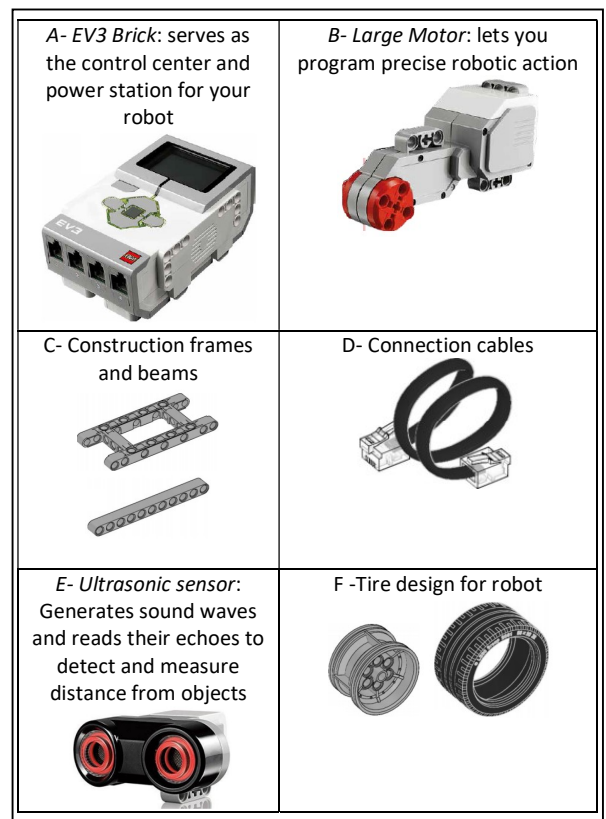

Figure 1: parts of LEGO robot.

The robot in Figure 4 was assembled for our experiment. During the assembly, we followed all necessary steps, as it was described by Carnegie Mellon's Robotics Academy manual [16]. Various robot constructions can be found on LEGO magazines and the design of the robot can depend on imagination of the designer [14].

Figure 4 is labeled with letters, the parts from Figure 1. Briefly, A- EV3 brick is control center, and block program from Figure 5 is uploaded to A through computer. B- large motors, is attached to A with cables, D, and gets directions from A on when to run motor or not. This motor will make robot move with help of the wheel – F.  So once the movement of the robot is arranged, it left to ultrasonic sensors, E, to sense the danger distance length decided by Figure 3, and send signal to A, to decide if robot should continue to move or further action is required. These further actions are presented in Figure 3 and Figure 5.

### 3.2 Hill Climbing

Figure 2 illustrates basically how the Hill climbing algorithm works, basically.

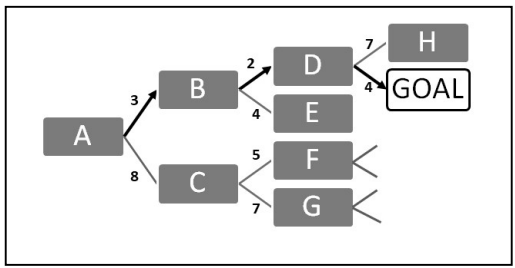

Figure 2: Searching for cheapest path

In our case, longer distance from wall is an optimal choice. Robot has 3 ultrasonic sensors (Figure 4); 1 in front- to sense obstacle in front, 2 by the sides, right left – they compare distance between the robot and the wall. Robot will go straight, when the gap is 10cm between robot and wall from front side, it will stop. Side sensors will compare distance to the respective walls, and robot will be ordered to continue to the side which has longer distance to the wall. Figure 3 illustrates the work of algorithm in robot. In this figure, robot needs to step back small from the wall, to take turn. We used word small, and this small distance we programmed 5cm length, but it may vary slightly, depending on robot position and ground friction.

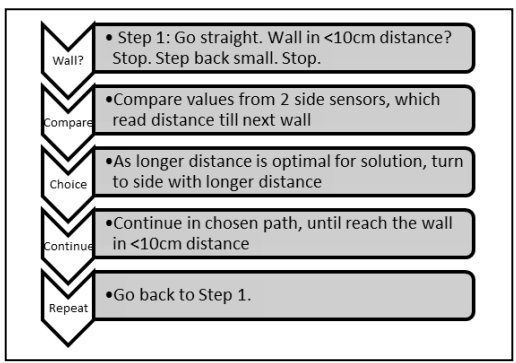

Figure 3: Decision steps using Hill Climb algorithm

In this way it will jump to the node which is giving better option. When it comes to exit from one line to another, it will evaluate it again longer distance to next wall and will follow it. In this way, it will try to choose best options to arrive to optimal solution.

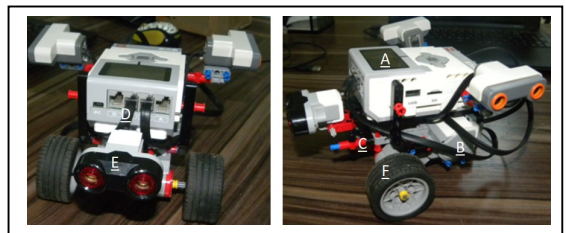

Figure 4: the assembled LEGO robot, ready for trial. Lettered parts are mentioned in Figure 1.

### 3.3 EV3 Programming

There are many ways to control EV3 robot, using well known languages; C, C++ and others. But to make it easier for education, LEGO preferred to develop a simple to use user interface, called Lab-VIEW [14].

This software can encode the robot using some simple blocks. At the back-end it uses C/C++ programming language. With object oriented programming, all directives are converted to some shapes, like in Figure 5. So user will just choose some blocks, to order robot what to do, and software will automatically program it in necessary way. Orders are basic, such as turn right or turn left, compare sensors etc. It is up to the user to program useful stuff so it can run smoothly. Figure 5 shows the implementation of algorithm, described in Figure 3, EV3 Lab View programming. Each part of the code is explained.

### 3.4 Maze design and construction

The definition of maze is a complicated system of paths or passages that people (robot in our case) try to find their way through, as it is described by Cambridge dictionary. Our maze is constructed from wood, with adjustable walls, so we can construct another maze. Size of board is proposed to be 120cm by 240cm. Road width will be 30cm. Height of walls 15cm to enable the robot sensor to adjust accordingly. Figure 6 shows how a sample maze will be constructed:

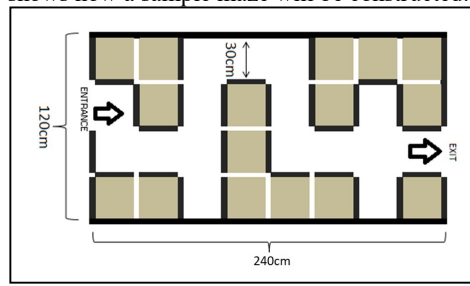

Figure 6: sample maze design

Complex designs are avoided as they can push robot for continuous loop, which need backtracking and cancellation of previous path, and this requires a different searching algorithm with more complex sensors. The main idea behind maze design was that robot should always have only 2 options to choose with one correct. Thus, in maze with intermediate level, finding longer path should bring you closer to solution.

For our experiment, we will try to use a maze with at least 2 right turns and 2 left turns, to understand if robot is choosing best option always. Figure 7 shows our real constructed maze for trial. Total length the robot is supposed to cover is 350cm

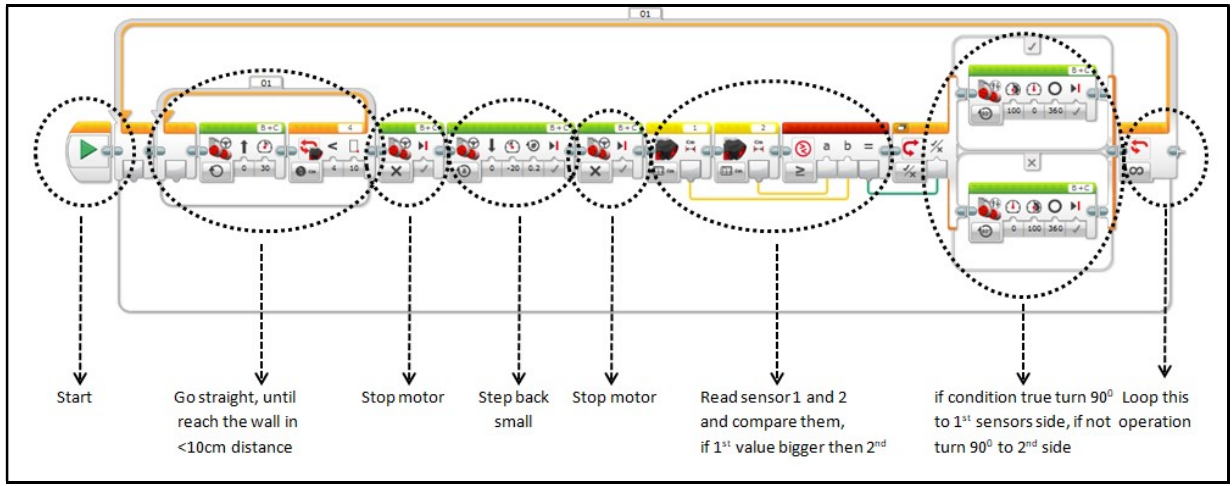

Figure 5: Lab view block programming of Hill climb design, described in Figure 3.

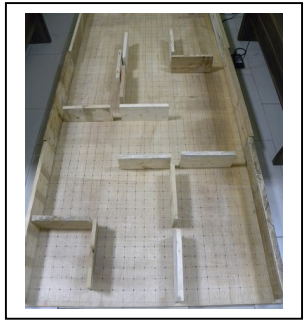

Figure 7: Wooden maze.

IV. RESULTS AND DISCUSSION

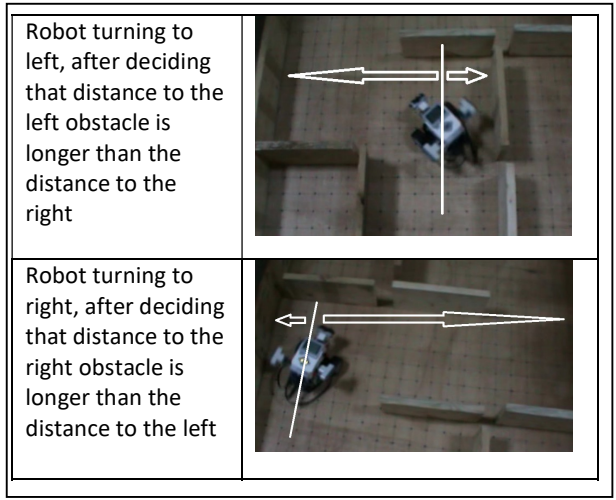

| | Robot turning to left, after deciding that distance to the left obstacle is longer than the distance to the right |
| | Robot turning to right, after deciding that distance to the right obstacle is longer than the distance to the left |

Figure 8: Pictures from trial

The total distance it covered was 3.5 meters, as we mentioned above. To help robot to make correct decisions, we programmed it to go slower. This can be done by adjusting speed of the robot during Lab View programming part, in Figure 5. Yet, number of turns is important as well. Table 1 shows results of first 4 trials. We didn't continue with trial as the fact that robot is traversing the maze was established.

| № | Result | Reason |
|---|---|---|
| Trial 1 | Failed at 2nd turn | Sensor misread |
| Trial 2 | Failed at 4th turn | Incomplete turn |
| Trial 3 | Successfully left maze | |
| Trial 4 | Successfully left maze | |

Table 1: Results of the trials

It turned 6 times in constructed maze. Total time needed to leave the maze was 37 seconds, in trial 3 and nearly same in trial 4. One can adjust it to go faster. But that is not our objective. Our main goal for this experiment was to see implementation of Hill climbing algorithm.

V. CONCLUSSION

In this paper, the authors assembled an EV3 robot and program it to use Hill climbing algorithm to traverse the maze, with complexity level of intermediate. We can conclude that it was successful one, as we achieved the aim, our robot traversed the maze and found exit.

In education, it is well proven fact that experimental approaches usually leading to better understanding of the concepts and theorems. Observing visually, how the robot is reacting, helps to understand the way the algorithm works, with its advantage and disadvantages.

Comparing our experiment to previous research trials, we aimed the same goal, to tackle the problem, in other words, to make the robot to complete given job successfully. But as we mentioned before, the outline of this experiment was to implement theory of searching algorithm. Once the algorithm steps are programmed into the robot, it is easy to observe the advantage and disadvantage of the algorithm.

We should underline the fact that the maze level was intermediate (relatively not hard), so was the implementation of the searching algorithm. In other words, the in maze construction, we avoided the conditions like death end, backtracking, and others. If one wants to solve maze with higher difficulty, some of solutions can be using better searching algorithm and more effective robot construction. For instance, our sensors were fixed, but the robot may have dynamic sensor reader, to tackle more difficult maze types. In same manner, an algorithm with backtracking options would help in various conditions.

## VI. ACKNOWLEDGEMENT

We would like to thank our supervisor Prof. Nwojo Agwu, Nile University of Nigeria, for his support and guidance during the experiment and paper preparation. The authors will also like to express appreciation to the anonymous reviewers whose comments helped to restructure the paper.